# A Connectionist Network Approach to Find Numerical Solutions of Diophantine Equations

Siby Abraham[1][*], Sugata Sanyal[2], Mukund Sanglikar[3]

**Abstract**— The paper introduces a connectionist network approach to find numerical solutions of Diophantine equations as an attempt to address the famous Hilbert's tenth problem. The proposed methodology uses a three layer feed forward neural network with back propagation as sequential learning procedure to find numerical solutions of a class of Diophantine equations. It uses a dynamically constructed network architecture where number of nodes in the input layer is chosen based on the number of variables in the equation. The powers of the given Diophantine equation are taken as input to the input layer. The training of the network starts with initial random integral weights. The weights are updated based on the back propagation of the error values at the output layer. The optimization of weights is augmented by adding a momentum factor into the network. The optimized weights of the connection between the input layer and the hidden layer are taken as numerical solution of the given Diophantine equation. The procedure is validated using different Diophantine Equations of different number of variables and different powers

——————————— ◆ — — — — — — — — —

## 1 INTRODUCTION

A Diophantine equation [1] [2] is a polynomial equation, given by

$$f(a_1, a_2, \ldots, a_n, x_1, x_2, \ldots, x_n) = N \qquad (1)$$

where $a_i$ and $N$ are integers. Diophantine equations (DE) have been named after the third century BC Alexandrian Mathematician Diophantus. In 'Arithmetic', he inquired about 150 algebraic problems and its solutions [3]. Such problems are now collectively referred as Diophantine equations. Diophantine equations and its particular types gave rise to some important results in Mathematics. Fermat's last theorem, Pythagorean triplets, Elliptic curves, Pell's equation and Waring Conjecture are few of them [4]. David Hilbert has been credited with giving a new direction to the interest shown on Diophantine equations by Mathematicians for centuries. In 1900 at the second International Congress on Mathematicians, he presented twenty three unsolved problems that he believed were important. The tenth problem he presented was: "Given a Diophantine equation with any number of unknowns and with rational integer coefficients: devise a process, which could determine by a finite number of operations whether the equation is solvable in rational integers." Since then, the problem is known as Hilbert's tenth problem [5] [6] and has been proved later that there exist no general method to solve this problem.

This paper is an attempt to find numerical solutions of Diophantine equations using Artificial Neural Networks or Connectionist Networks [7] [8] [9]. The paper introduces a feed forward three layer neural network with back propagation learning [10] [11]. It experimentally validates its effectiveness by finding numerical solutions of a class of Diophantine equations given by equation (2).

$$x_1^{p_1} + x_2^{p_2} + \ldots + x_n^{p_n} = N \qquad (2)$$

The paper is structured in six sections. Section 2 gives an overview of Connectionist Networks. Section 3 describes the related works done in the field of Diophantine equations and back propagation networks. Section 4 explains the procedure proposed to find numerical solutions of the given class of Diophantine equations. Section 5 deals with implementation and experimental results. Section 6 discusses final conclusion.

## 2 CONNECTIONIST NETWORKS

Connectionist network, which are also called Artificial Neural Networks (ANN), is a computational model based on the working of biological neural network of brain. ANN resembles brain in two respects [12]: Knowledge is acquired by the network through a learning process and inter-neuron connection strengths known as synaptic weights are used to store the knowledge. Connectionist networks are comprised of densely interconnected adaptive processing units, known as neurons [13], which help to replace the usual 'programming' in solving problems by the concept of 'learning by example'. This is made possible using many neural network models and learning algorithms.

There are different neural network models. They can be broadly classified as Feed forward Network, Radial basis Function Network, Kohonen Self Organizing Network, Recurrent Network, Stochastic Network and Modular network [14]. These network models use different learning algorithms for training the network. These algorithms come under three

———————————————
[1][*]Dept of Mathematics & Statistics, G.N.Khalsa College, University of Mumbai, India. Email: sibyam@gmail.com
[2]Corporate Technology Office, Tata Consultancy Services, Gateway Park, Andheri (E), Mumbai, India. Email: sugata.sanyal@tcs.com.
[3]Dept of Mathematics, Mithibai College, University of Mumbai,

different learning paradigms. They are supervised learning, un-supervised learning and reinforcement learning.

Back Propagation (BP) learning algorithm, which is also called generalized delta rule, is a supervised learning algorithm for a feed forward multi layer neural network and is one of the most widely used learning processes in ANN. The basic architecture of the Back Propagation Network [15] [16] [17] consists of an input layer, one output layer and one or more intermediate layers known as hidden layers. The training algorithm of back propagation involves four stages. They are: initialization of weights, feed forward, back propagation of errors and updating of the weights. The name 'back propagation' originates from the fact that the errors of hidden units are derived from propagating backward the error obtained from the output unit. After a sufficient number of times, the error values will be minimized to the expected level and the output generated will be almost matching with the expected result. This process is repeated for different input-output pairs and the same helps to obtain the optimized weights finally. Now, the system is said to have been trained or learnt and hence is expected to classify similar input values and give predictable output values in real time.

## 3 RELATED WORKS

There have been many attempts to unravel the mysteries around Hilbert's tenth problem. Davis et al. [18] proved that an algorithm to determine the solvability of all exponential Diophantine equations is impossible. Matiyasevich [19] extended this work by showing that there is no algorithm for determining whether an arbitrary Diophantine equation has integral solutions, ending attempts of centuries for finding a general method.

Though Hilbert's famous tenth problem was settled unilaterally and conclusively, Diophantine Equations, which has more open problems than results [20], is an area which is engaging and challenging. This has become all the more important as the significance of Diophantine equations has not just been restricted to the abstract and theoretical realm of Mathematics as it has found newer application areas in the fields like Public key cryptosystems [21] [22], Computable economics [23], Elliptic curves [24] [25], Data dependency in super computers [26] [27] and Theoretical Computer Science [28] [29].

In this context, attempt to find numerical solutions of DE is inspiring and interesting. However, this turned out to be a hard problem as the search space [30] [31] of (1) consists of $N^n$ elements. Since following an exhaustive, gradual and incremental method involved high degree of computational complexity, soft computing approach of Artificial Intelligence [32] [33] were looked at. There have been some attempts to find numerical solutions of Diophantine Equations in this direction. Abraham and Sanglikar [34] applied Genetic Algorithm (GA) with mutation and crossover as genetic operators to find numerical solutions of a class of DE. They [35] explained the process of avoiding premature converging points using 'Host Parasite Co-evolution' in a typical GA. Abraham and Sanglikar [36] also explained a method involving evolutionary and co-evolutionary computing techniques to find numerical solution of DE. Their paper [37] described a simulated annealing based computational approach to find numerical solution of a DE. Abraham et al [38] introduced particle swarm optimization for finding numerical solutions of such equations.

Though there are some attempts to find numerical solutions of Diophantine equations using various soft computing paradigms [39], to where the Connectionist networks belongs, the authors are yet to find literature on Diophantine equations using connectionist networks except the work by Joya et al [40]. They have showed that higher order Hopfield networks can be used to find numerical solutions of DE. It is quite surprising that there is only one work of connectionist networks in the field of Diophantine equations as connectionist networks with learning mechanisms like back propagation are tried and tested mechanisms on a wide range of problems belonging to diverse backgrounds, few of which are mentioned in the following paragraph. It is all the more relevant as the back propagation is the often used learning mechanism of all others in the connectionist networks. This paper is an honest and sincere attempt to fill that void.

Cun et al [41] described an application of back propagation networks to recognize handwritten digits of low level presentation of data with minimal preprocessing. Kamruzzaman et al [42] introduced a double back propagation assisted neural network for character recognition which first preprocesses input pattern to produce a translation, rotation and scale invariant representation and then classifies characters using a neural net classifier. Temdee et al [43] explained back propagation based face recognition where fractal codes from the edge pattern of the face region are fed as inputs to a back propagation neural network for training the network and hence identifying a person. Durai and Anna Saro [44] described how back propagation network can be used for image compression wherein the gray levels of the pixels in an image and their neighbors are mapped such a way that the difference in the gray levels of the neighbors with the pixel is minimum so that the compression ratio as well as the convergence of the network is improved. R Chang et al [45] dealt with a learning methodology using back propagation neural networks with sample-query and attribute-query in developing an intrusion detection system (IDS). Hossam Osman et al [46] proposed a back propagation neural network for the classification of ship targets in airborne synthetic aperture radar (SAR) imagery. K Shihab [47] talked about a symmetric encryption mechanism based on artificial neural networks where a multi-layer neural network, which is trained by back propagation learning algorithm, is used for the decryption scheme and the public key creation. Pokorny and Smizansky [48] proposed a method called Page Content Rank (PCR) based on the page content exploration in Web Content Mining.

## 4 PROPOSED SYSTEM

The system developed to find numerical solutions of Diophantine equations, which we call Neural Network based Diophantine Equation Solver (NEURO-DOES) introduces a dynamically constructed architecture for the feed forward neural network with back propagation learning.

### 4.1 Network architecture
NEURO-DOES uses network architecture of three layers of one each of input, hidden and output layers. The number of neurons in the hidden layer and output layer is fixed to be one. The number of neurons in the input layer changes corresponding to the type of the equation. If the given Diophantine equation has n variables then input layer also will have that many numbers of neurons. The powers of the given Diophantine equation are given as input to the neurons in the input layer. The values of the variables are taken as weights of the connection between the input layer and the hidden layer. The learning process starts by assigning small random integral values as initial weights. The solution of the given Diophantine equation is the resultant weights of the connections between the input layer and the output layer obtained once the system is fully trained.

### 4.2 Initial weights

All connection weights, which are initialized as small random integral values, are taken to be positive integers as we are interested in only positive integral solutions of the equation. Initial weights for the connection between the input layer and the hidden layer are chosen randomly, as is the usual practice, from a small range between 1 and 10. The rare chance of the procedure not converging to the optimum weights and hence not getting the solution for the given Diophantine equation is resolved by having the option to restart the whole process with initial random distribution of weights from another range. The weight of the connection between the hidden layer and the output layer is assigned to be one and is kept fixed.

### 4.3 Feed forward

Each neuron in the input layer receives the powers pi of the variables as input signal and transmits them into the next layer i.e. hidden layer. The weights $w_i$ connecting the neurons in the input layer and the hidden layer correspond to the prospective values of the variables, which satisfies the equation. The units in the hidden layer receive signals in the form of $w_i{}^{pi}$, which differs from $w_i * pi$ as is the usual practice in Back propagation networks, to accommodate the constraints unique to the Mathematical problem under study. The neuron in the hidden layer calculates the exponentially weighted sum of the signals. Thus, we have

$$Q = \Sigma\ w_i{}^{pi} \qquad (3)$$

as the output from the hidden unit. The unit in the hidden layer is connected to the unit in the output layer. The weight of this connection is fixed to be 1. Here, we have only one connection to the output layer: Q with weight 1, which is a pseudo weight, as the value of this weight does not have any bearing on the final result and is fixed. The unit in the output layer is taken as the weighted sum of this signal and output as a linear function. This is the output of the output unit and is denoted by OP. If OP is equal to N, the network is said to have been trained to give optimized weight values for the given input and these optimized values are taken as the solution of the given equation. Any other value of OP shows an error and requires further training of the network.

### 4.4 Back Propagation of Errors

The value of OP other than N shows that there is an error in the output and the system needs to be trained. The value of this error is calculated as

$$E = Error = OP - N. \qquad (4)$$

E can either be positive or be negative.
We define
$$\delta = Sgn\ (E) = 1\ \text{if}\ E > 0$$
$$= -1\ \text{if}\ E < 0$$
$$= 0\ \text{if}\ E = 0. \qquad (5)$$

This value of '$\delta$' is propagated back to the unit in the hidden layer. We call '$\delta$' as the error factor. Since there is only one connection connecting hidden layer and output layer, the error factor of the previous layer is also taken as '$\delta$'. In practice, the value of the error factor as +1 means that the network needs to increment the weights of the connections between input layer and hidden layer. Similarly the –1 value directs the network to decrement the weights between the input layer and hidden layer. When '$\delta$' takes value 0, then the system is said to have been learnt and the optimized weights give the solution. This back propagation of +1 or –1 error message helps the network to optimize the weight vector of the connection between the input layer and hidden layer.

### 4.5 Update of Weights

The amount by which the network weights between the input layer and the hidden layer are to be incremented or decremented is decided by the extent of the error. This quantity is taken to be dependent on |Error| 1/ (Max * n), where 'Max' is the maximum of the powers of the variables and 'n' is the number of variables of the given equation. The weights of the connection between the input layer and the hidden layer are updated by $\Delta$ wi based on the following three different cases:

Case 1. $\delta > 0$: If $\delta > 0$ the weight updating is done in the straight forward fashion using formula:

$$\Delta wi = \delta\ |Error|\ 1/(Max * n), \text{for}\ i = 1, 2, \ldots, n \qquad (6)$$

Equation (6) helps weight updating faster. This happens usually at the beginning of the procedure when the weights are small random values and the error is quite high. The quick pace of the weight updating facilitates the convergence of the weights to the vicinity of the optimized weights.

Case 2. $\delta < 0$ and $w_i \neq 1$: If $\delta < 0$ and if $w_i \neq 1$ for all i=1, 2… n then the weights are adjusted using the formula given by

$$\Delta w1 = \delta.\min\ \{1,\ |Error|\ 1/\ (Max * n)\} \qquad (7)$$

$$\text{and}\ \Delta w_i = 0\ \text{for}\ I = 2,\ \ldots\ N. \qquad (8)$$

Equation (7) helps to adjust the weight of the first connection by keeping other connection weights constant. The selection of nominal adjustment of only one connection is deliberate as otherwise weights might oscillate between the optimized weights in the weight space. This way the convergence of the weight is directed by focusing on one weight at a time when the error value crosses the positive boundary.

Case 3. $\delta < 0$ and $wj = 1$: If $\delta < 0$ and if $wj = 1$ for some $j = i$ where $i= 1, 2…n$, then weight adjustment is done using the following formulae:

$$\Delta w_j = count * \min \{1, |Error| \cdot 1/ (Max * n)\} \quad (9)$$
$$\Delta w_{j+1} = \delta \cdot \min \{1, |Error| \cdot 1/ (Max * n)\} \quad (10)$$
$$\text{and } \Delta w_i = 0 \text{ for } i \neq j, j+1 \quad (11)$$

Here 'count' is the number of previous consecutive iterations with $\delta <0$. This case arises because the reduction of weights cannot be applied when the weight of any connection hits 1, the smallest positive integer, as we are interested only in the positive integral solution of the given equation. This necessitates a modification of Equation (7) corresponding to the case 2. This is achieved by incorporating Equations (9), (10) and (11) in the process. The case is handled in three folds. Firstly, the weight, whose value is 1, is adjusted using a momentum as shown in Eq. (9). Momentum is added to a neural network to speed up the convergence in the positive direction. Here the quantity 'count' is taken as the momentum factor for speedier convergence. Secondly, the weight of the connection, next to the connection with weight 1, is reduced nominally as provided by Eq. (10). Thirdly, all other connection weights are kept constant as shown by Eq. (11). In this way the procedure is directed towards the optimal weights.

## 5 EXPERIMENTAL RESULTS

The procedure discussed in the work is implemented in C-language. The data structures like arrays, pointers and structures are used to implement the program constructs. Experimental results of running the program for different cases show encouraging results.

### 5.1 Results on equations with different variables

Table 1 shows the results obtained when the system was run on Diophantine equations with varying number of variables. It discusses solutions of nine different Diophantine equations with nine different number of variables, changing from two to ten. The system provides solutions even when the number of variables is competitively high. The last column of the table 1 shows that if the number of variables is comparatively less, then the system gives the solutions in much lesser iterations as the convergence of the optimized weights takes place very fast.

**Table 1** Results on equations with varying number of variables

| Sr. No | Diophantine Equation | No. of variables | Solution Found | Iterations required |
|---|---|---|---|---|
| 1 | $x_1^2+x_2^2 = 149$ | 2 | 10, 7 | 10 |
| 2 | $x_1^2+x_2^2+x_3^2 = 244$ | 3 | 12, 6, 8 | 6 |
| 3 | $x_1^2+x_2^2+…+x_4^2= 295$ | 4 | 1, 2, 13, 11 | 51 |
| 4 | $x_1^2+x_2^2+….+x_5^2= 325$ | 5 | 1, 5, 9, 7, 13 | 33 |
| 5 | $x_1^2+x_2^2+….+x_6^2= 420$ | 6 | 1, 1, 2, 7, 13, 14 | 97 |
| 6 | $x_1^2+x_2^2+….+x_7^2= 450$ | 7 | 1, 1, 2, 2, 10, 14, 12 | 457 |
| 7 | $x_1^2+x_2^2+….+x_8^2= 590$ | 8 | 1, 1, 1, 1, 5, 13, 14, 14 | 1669 |
| 8 | $x_1^2+x_2^2+….+x_9^2= 720$ | 9 | 1, 1, 1, 2, 2, 14, 12, 12, 15 | 1373 |
| 9 | $x_1^2+x_2^2+….+x_{10}^2=956$ | 10 | 1, 1, 1, 1, 1, 1, 1, 15, 20, 18 | 9068 |

### 5.2 Results on equations with varying degrees

Table 2 demonstrates the effectiveness of the system by showing the results obtained for different values of powers. It also has nine different Diophantine equations but the powers of the equations are varied this time. Different powers from two to ten are used with different equations. It shows that the system gives the numerical solutions even when powers are large.

On comparing the last columns of table 1 and table 2, it is observed that the number of iterations required to find the solutions is much lesser in the case of powers than in the case of different variables. This is because the search space is not very complex in the case of large powers, which is not the case for equations with large number of variables.

Table 2  Results on equations with varying degrees

| Sr. No | Diophantine Equation | Degree | Solution | Iterations required |
|---|---|---|---|---|
| 1 | $x_1^2 + x_2^2 = 625$ | 2 | 20, 15 | 7 |
| 2 | $x_1^3 + x_2^3 = 1008$ | 3 | 2, 10 | 23 |
| 3 | $x_1^4 + x_2^4 = 1921$ | 4 | 6, 5 | 10 |
| 4 | $x_1^5 + x_2^5 = 19932$ | 5 | 7, 5 | 9 |
| 5 | $x_1^6 + x_2^6 = 47385$ | 6 | 6, 3 | 6 |
| 6 | $x_1^7 + x_2^7 = 4799353$ | 7 | 9, 4 | 5 |
| 7 | $x_1^8 + x_2^8 = 16777472$ | 8 | 8, 2 | 2 |
| 8 | $x_1^9 + x_2^9 = 1000019683$ | 9 | 3, 10 | 12 |
| 9 | $x_1^{10} + x_2^{10} = 1356217073$ | 10 | 7, 8 | 4 |

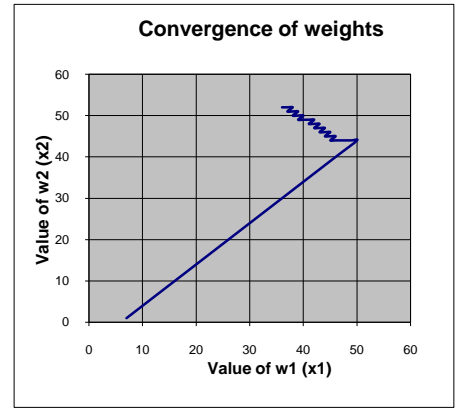

Figure 1 Convergence of weights

### 5.3 RESULTS ON EQUATIONS WITH VARYING DEGREES

Figure 1 shows convergence of weights in the weight space for an elementary equation given by

$$x_1^2 + x_2^2 = 4000 \qquad (12)$$

Initially, the convergence is very rapid and the process follows an almost linear pace. This is because of the higher learning rates at the initial stage of the convergence. As the weights reaches in the vicinity of optimized weights, the procedure slows down as the error becomes comparatively small. Now on, the learning rates are reduced and the process gradually moves towards the optimum weight values.

### 5.4 REDUCTION OF ERROR

Figure 2 shows the reduction of errors corresponding to the equation (12) on the process of optimizing the weights. It shows that there are large error values at the initial stage of the weight optimization process. Gradually, they are brought to small values as the process becomes matured. Toward the optimum weights, the error values fluctuate between positive and negative values within a smaller range and are brought to the optimum value by the momentum factor incorporated in the procedure. The contribution of momentum factors to fasten the optimization process is evident from the figure.

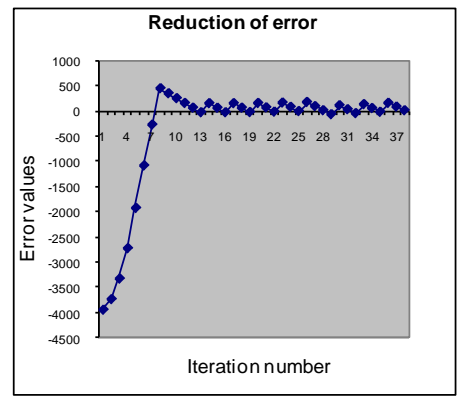

Figure 2 Reduction of error

### 5.5 LEARNING RATES

Figure 3 shows the learning rates attained by the procedure during the process of optimizing weights for the same equation (12). Initially, there is a comparatively large learning rate. But as the process becomes stabilized, the learning reduces and the changes in two consecutive iterations become negligible. The gradual convergence to the optimized weights is represented by the almost steady learning rates at those stages. The figure also shows the iteration number where the learning rate becomes zero, which shows that the solution has been encountered by the system.

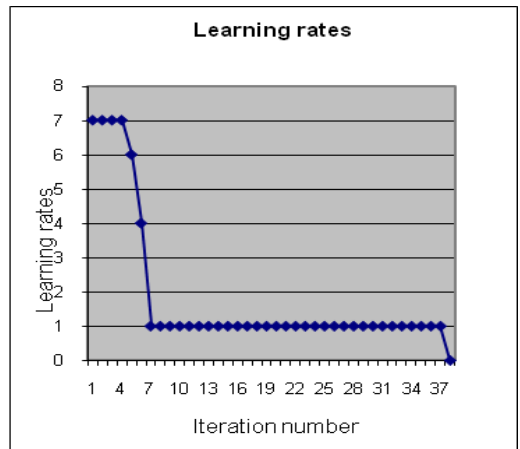
Figure 3 the learning rates attained by the procedure

# 6 CONCLUSIONS AND FUTURE WORK

The paper introduces a connectionist network based approach to find numerical solutions of Diophantine equations. It uses a three layer feed forward network with back propagations as the sequential learning strategy. The judicious selection of network architecture to take care of the domain specificity of Diophantine equations resulted in the convergence of optimum weights giving numerical solutions of the equations.

Though the system could offer solutions of Diophantine equations with different number of variables and different powers, the procedure has the tendency to find solutions of which coordinates are more close to one another. The further works involve obtaining solutions with distinct coordinates as a Diophantine equation can have many solutions. This is expected to be achieved by a better learning strategy.